\documentclass[preprint,12pt]{elsarticle}

\usepackage[margin=0.80in]{geometry}
\usepackage{float}
\usepackage{adjustbox}
\usepackage{wrapfig}
\usepackage{soul}
\usepackage{footnote}
\usepackage{lineno}
\usepackage{hyperref}
\usepackage{listings}
\usepackage{graphicx}

\lstdefinestyle{R}{
    language        = R,
    frame           = lines, 
    basicstyle      = \footnotesize,
    keywordstyle    = \color{black},
    stringstyle     = \color{black},
    commentstyle    = \color{black}\ttfamily
}

\makeatletter
\def\ps@pprintTitle{%
 \let\@oddhead\@empty
 \let\@evenhead\@empty
 \def\@oddfoot{\centerline{\thepage}}%
 \let\@evenfoot\@oddfoot}
\makeatother

\usepackage{multirow}
\usepackage{graphicx}
\usepackage{color}
\usepackage{pdfpages}

\usepackage{caption}
\usepackage{subcaption}
\usepackage{amssymb,amstext,amsmath}
\usepackage{float}
\usepackage{booktabs}
\usepackage{array,ragged2e}
\usepackage[table]{xcolor}

\usepackage{multirow}
\usepackage{algorithm}
\usepackage{algpseudocode}
\usepackage{pifont}

\biboptions{sort&compress}
\begin{document}
\begin{frontmatter}

\title{A Multi-Scale Isolation Forest Approach for Real-Time Detection and Filtering of FGSM Adversarial Attacks in Video Streams of Autonomous Vehicles}

\author[MCSaddress]{Richard Abhulimhen}
\ead{r.abhulimhen41@my.benedict.edu}

\author[MCSaddress]{Negash Begashaw}
\ead{negash.begashaw@benedict.edu}

\author[NCATaddress]{Gurcan Comert}
\ead{gcomert@ncat.edu}

\author[CLEMaddress]{Chunheng Zhao}
\ead{chunhez@g.clemson.edu}

\author[CLEMaddress]{Pierluigi Pisu}
\ead{pisup@clemson.edu}

\address[MCSaddress]{Computer Science and Engineering Department, Benedict College, Columbia, SC 29204 USA}
\address[NCATaddress]{Comp. Data Science and Eng., North Carolina A\&T State University, Greensboro, NC 27411 USA}
\address[CLEMaddress]{Department of Automotive Engineering, Clemson University, Clemson, SC 29607 USA}

\begin{abstract}
Deep Neural Networks (DNNs) have demonstrated remarkable success across a wide range of tasks, particularly in fields such as image classification. However, DNNs are highly susceptible to adversarial attacks, where subtle perturbations are introduced to input images, leading to erroneous model outputs. In today's digital era, ensuring the security and integrity of images processed by DNNs is of critical importance. One of the most prominent adversarial attack methods is the Fast Gradient Sign Method (FGSM), which perturbs images in the direction of the loss gradient to deceive the model. 

This paper presents a novel approach for detecting and filtering FGSM adversarial attacks in image processing tasks. Our proposed method evaluates 10,000 images, each subjected to five different levels of perturbation, characterized by $\epsilon$ values of 0.01, 0.02, 0.05, 0.1, and 0.2. These perturbations are applied in the direction of the loss gradient. We demonstrate that our approach effectively filters adversarially perturbed images, mitigating the impact of FGSM attacks. 

The method is implemented in Python, and the source code is publicly available on GitHub for reproducibility and further research.
\end{abstract}


\end{frontmatter}

\section{Introduction}

\noindent Deep Neural Networks (DNNs) have revolutionized the field of image classification, achieving unprecedented accuracy across a wide array of applications. From autonomous vehicles and medical imaging to facial recognition and content moderation, DNNs have become indispensable to many critical technological infrastructures. However, this widespread adoption has exposed a significant vulnerability: DNNs are susceptible to adversarial attacks. These attacks involve subtle, often imperceptible, manipulations of input data that mislead neural networks, posing a serious threat to the reliability and security of AI systems in real-world scenarios.

Among the various adversarial attack methods, the Fast Gradient Sign Method (FGSM) has emerged as a particularly potent and extensively studied technique. Introduced by Goodfellow et al. in 2014, FGSM exploits the linearity of many machine learning models in high-dimensional spaces. The attack works by calculating the gradient of the loss function with respect to the input image and then perturbing the image in the direction that maximizes this loss. What makes FGSM especially concerning is its ability to generate adversarial examples quickly and with minimal computational resources, often resulting in perturbations that are imperceptible to the human eye yet cause severe misclassifications by the target model.

\section{Research Background}

\noindent The potential consequences of such attacks are far-reaching and can be catastrophic. In autonomous vehicles, for example, an adversarial attack could cause a car to misinterpret a stop sign as a speed limit sign, leading to dangerous traffic situations. In medical imaging, adversarial perturbations could lead to incorrect diagnoses with life-threatening consequences. Even in less critical applications, such as content moderation on social media, adversarial attacks could bypass filters, exposing users to harmful or illegal content.

Given these risks, the development of robust defense mechanisms against adversarial attacks, particularly FGSM, has become a crucial area of research in the machine learning community. Our work contributes to this field by presenting a novel method for filtering adversarial images generated using FGSM. We evaluated our approach using a dataset of 10,000 diverse images and applied five distinct levels of perturbation, characterized by $\epsilon$ values of 0.01, 0.02, 0.05, 0.1, and 0.2. This allowed us to assess the effectiveness of our filtering method across a spectrum of attack intensities, from subtle perturbations that evade detection to more aggressive alterations.

The research landscape on adversarial attack detection and mitigation is rich and varied. A. Higashi et al. \cite{Higashietal} investigated noise removal operations and proposed a simple yet effective adversarial image detector. Their work underscores the potential of traditional image processing techniques in combating adversarial attacks. Liang et al. \cite{Liangetal} approached adversarial perturbations as a form of noise, employing scalar quantization and spatial smoothing filters to mitigate their impact. This approach highlights the viability of treating adversarial attacks as a noise reduction problem, opening the door to a wide array of signal processing techniques for defense.

Similarly, Vadim Ziyadinov and Maxim Tereshonok \cite{Vadimetal} addressed the challenge of high-frequency noise in Convolutional Neural Networks (CNNs), showing that low-pass filtering can significantly improve recognition accuracy under adversarial conditions. Their work emphasizes the importance of understanding the frequency domain characteristics of adversarial perturbations. Fei Wu et al. \cite{Feietal} introduced an innovative method that combines noise injection with adaptive Wiener filtering, showcasing a novel defense mechanism where controlled noise addition mitigates adversarial effects.

Wei Kong et al. \cite{Weietal} proposed a comprehensive threat model and evaluation framework that standardizes the assessment of adversarial defenses, providing a benchmark for future research. Kishor Datta Gupta et al. \cite{Kishoretal} developed a dynamic framework that generates random image processing sequences at testing time, introducing unpredictability that complicates the attacker's ability to craft effective adversarial examples.

Our work builds upon these foundational studies, introducing a filtering method that combines elements of anomaly detection, image processing, and machine learning to create a more robust defense against FGSM attacks. By integrating and extending prior techniques, we offer a generalizable solution that enhances the resilience of image classification models in the face of adversarial threats.

\section{Research Objectives}

\noindent This paper aims to address the challenges posed by FGSM adversarial attacks by developing a novel filtering method. The objectives include:
\begin{itemize}
    \item Developing a robust mechanism for detecting adversarial images generated by FGSM.
    \item Evaluating the method's effectiveness across various levels of perturbation intensity.
\end{itemize}

\section{Scope of Research}

To ensure the robustness and generalizability of our method, we utilized a diverse dataset comprising 10,000 original images. These images were carefully selected to represent various real-world scenarios, including natural scenes, objects, and abstract patterns. To thoroughly evaluate the effectiveness of our filtering method, we applied five distinct levels of FGSM perturbations to each image, resulting in a dataset of 60,000 images (10,000 original + 50,000 perturbed).

The perturbation levels were controlled by the parameter $\epsilon$, set at values of 0.01, 0.02, 0.05, 0.1, and 0.2. These $\epsilon$ values represent a range of attack intensities, from subtle perturbations that might evade human detection to more aggressive distortions that are easier to identify. By applying these perturbations in the direction of the loss gradient, we simulated a variety of adversarial attack scenarios, allowing us to evaluate our method’s performance across different attack intensities.

\section{Actions Performed}

\noindent This paper systematically addresses the problem through the following actions:

\begin{itemize}
    \item \textit{Implementation Steps} \ref{implementation-steps} outlines the structured, multi-stage implementation methodology:
    \begin{itemize}
        \item \textit{Image Conversion and Storage} (\ref{image-conversion-and-storage}) explains the process of image conversion and storage, detailing the extraction of frames from video and the creation of a structured dataset.
        \item \textit{Adversarial Attack Generation} (\ref{adversarial-attack-generation}) elaborates on the generation of adversarial images using the FGSM algorithm, simulating attacks at various perturbation levels.
        \item \textit{Detection of Adversarial Images} (\ref{detection-of-adversarial-images}) describes the detection of adversarial images through a dual-pronged approach utilizing One-Class SVM and Isolation Forest for robust anomaly identification.
        \item \textit{Visualization and Decoration} (\ref{visualization-and-decoration}) focuses on the visualization and annotation of detection results, including anomaly score plotting, decision boundary visualizations, and image decoration with intuitive color-coded borders.
    \end{itemize}
    \item \textit{Method Summary} (\ref{method-summary}) provides a comprehensive summary of the proposed method, integrating all key processes to present a robust and scalable approach for adversarial image detection.
\end{itemize}

\subsection{Implementation Steps}
\label{implementation-steps}
Our implementation follows a structured, multi-stage approach aimed at efficiently processing, analyzing, and filtering adversarial images:

\begin{figure}[H]
    \centering
    \includegraphics[width=0.38\textwidth]{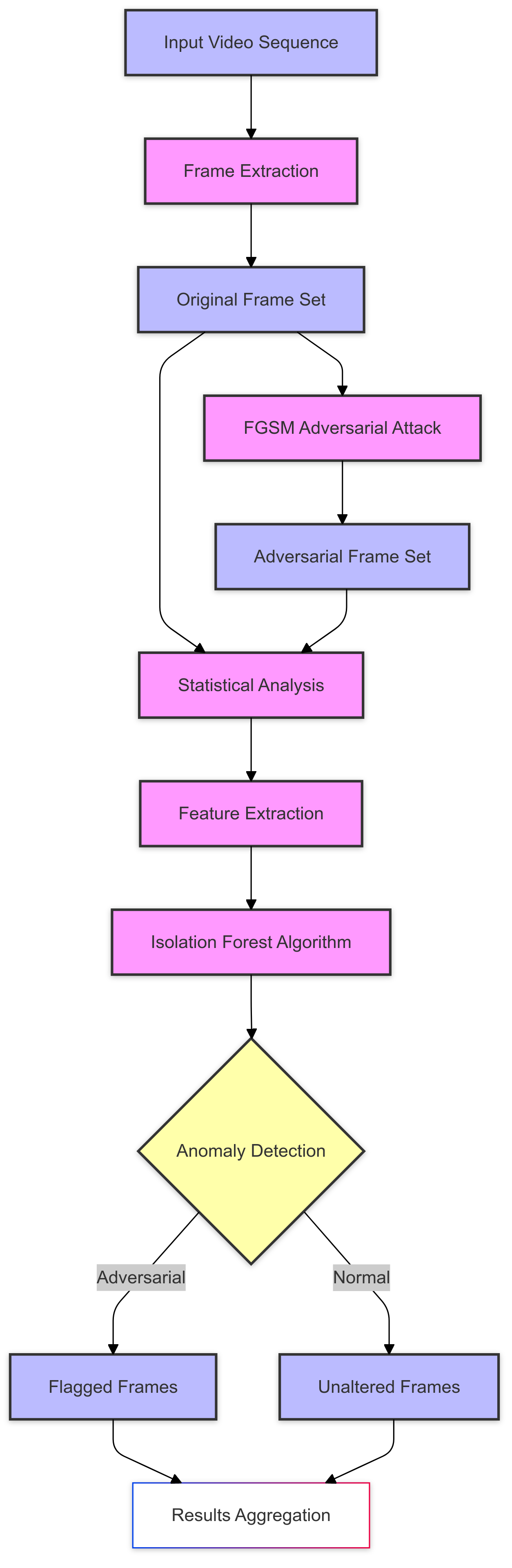}
    
    \caption{Adversarial Image Detection Pipeline. The flowchart illustrates the step-by-step process of detecting adversarial images, starting from video input and frame extraction to final classification and aggregation. Processes (pink), data (light blue), and decision points (yellow) are color-coded for clarity.}
    \label{fig:adversarial_pipeline}
\end{figure}

\subsubsection{Image Conversion and Storage}
\label{image-conversion-and-storage}
The first step in our pipeline involves converting the original video content into individual frames. This step is crucial for enabling frame-by-frame analysis and adversarial attack simulation. Each extracted frame is systematically stored in a designated output directory, creating a structured dataset that forms the basis for subsequent processing.

Once the frames are extracted, each image undergoes FGSM attacks at the predetermined perturbation levels. This generates a comprehensive set of adversarial images, each reflecting a different intensity of attack. These perturbed images are stored in a separate directory, ensuring a clear distinction between original and adversarial content.

\subsubsection{Adversarial Attack Generation}
\label{adversarial-attack-generation}
The FGSM attack is implemented meticulously to accurately represent real-world adversarial attack scenarios. For each original image, the perturbations are calculated and applied using the FGSM algorithm, defined as follows:
\[
\text{perturbed image} = x + \epsilon \cdot \text{sign}(\nabla_x J(\theta, x, y))
\]
where $x$ represents the original image, $\epsilon$ is the perturbation magnitude, $J$ is the loss function, $\theta$ represents the model parameters, and $y$ is the true label. This process is repeated for each $\epsilon$ value, producing a diverse set of adversarial examples for each original image.

The implementation utilizes a pre-trained ResNet-50 model as the target neural network, leveraging its deep architecture for gradient computation. For each frame, the image is first preprocessed to match ResNet's input requirements, including normalization and resizing to 224×224 pixels. The attack computes the gradient of the loss with respect to the input image using PyTorch's autograd functionality, which efficiently tracks the computational graph for gradient calculation.

The loss function $J$ is implemented as the cross-entropy loss between the model's predictions and the original class labels. By using the sign of the gradient rather than its raw values, FGSM efficiently generates adversarial perturbations that maximize the loss in the direction of each pixel's gradient, while maintaining a bounded perturbation magnitude controlled by $\epsilon$. This approach ensures that the perturbations remain imperceptible while effectively misleading the model's predictions.

The attack preserves the image's structural integrity by applying clipping to ensure all pixel values remain within the valid range [0,1] after perturbation. This maintains the visual quality of the frames while achieving the desired adversarial effect. The perturbation process is optimized to operate efficiently on video frames, allowing for real-time generation of adversarial examples while maintaining the temporal consistency of the video sequence.

\begin{figure}[H]
    \centering
    
    \begin{subfigure}{0.3\textwidth}
        \centering
        \includegraphics[width=\textwidth]{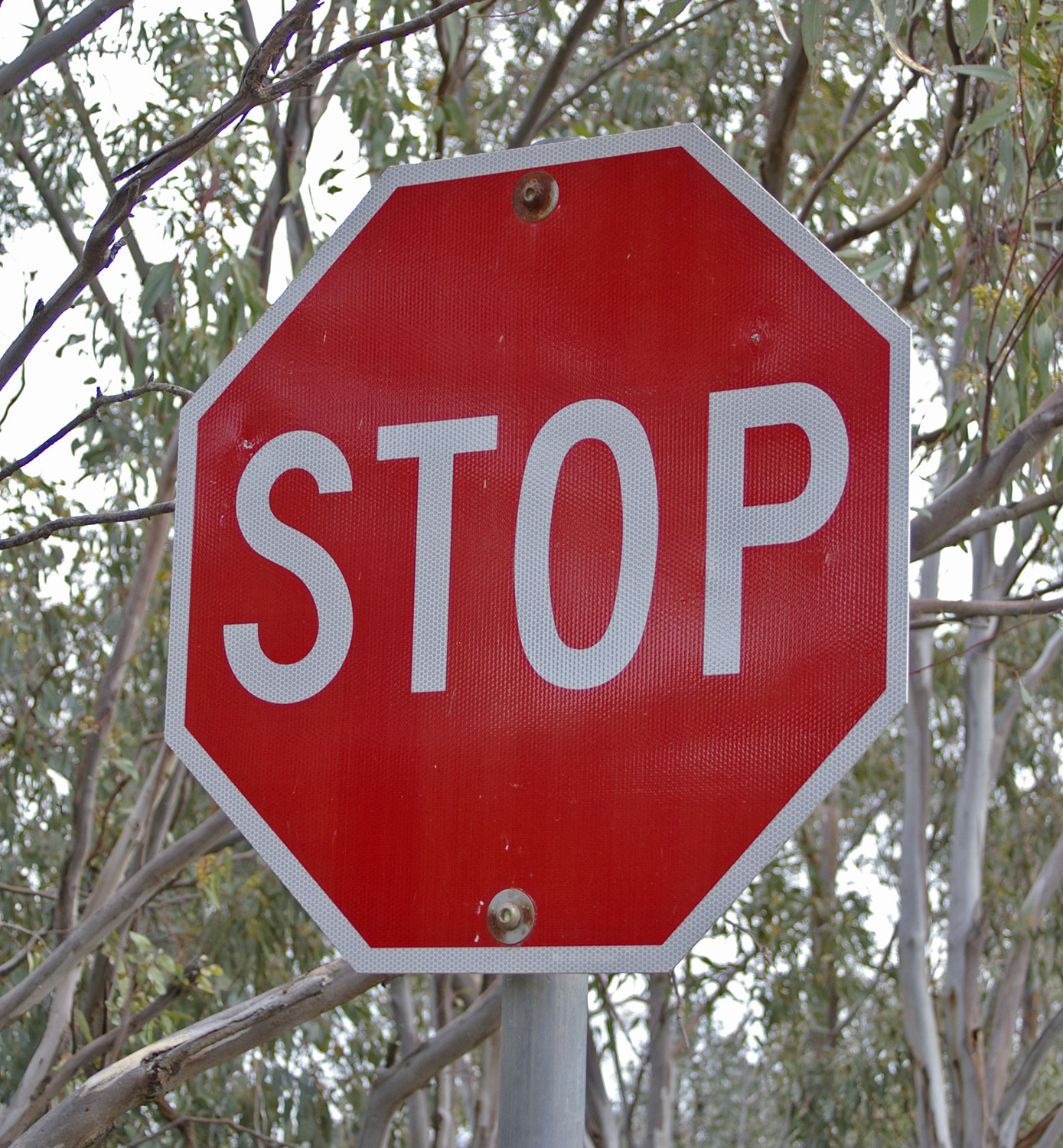}
        \caption{Original unperturbed image showing a clear, natural representation of the input sample.}
        \label{fig:original}
    \end{subfigure}
    
    \vspace{1em}
    
    \begin{subfigure}{0.31\textwidth}
        \centering
        \includegraphics[width=\textwidth]{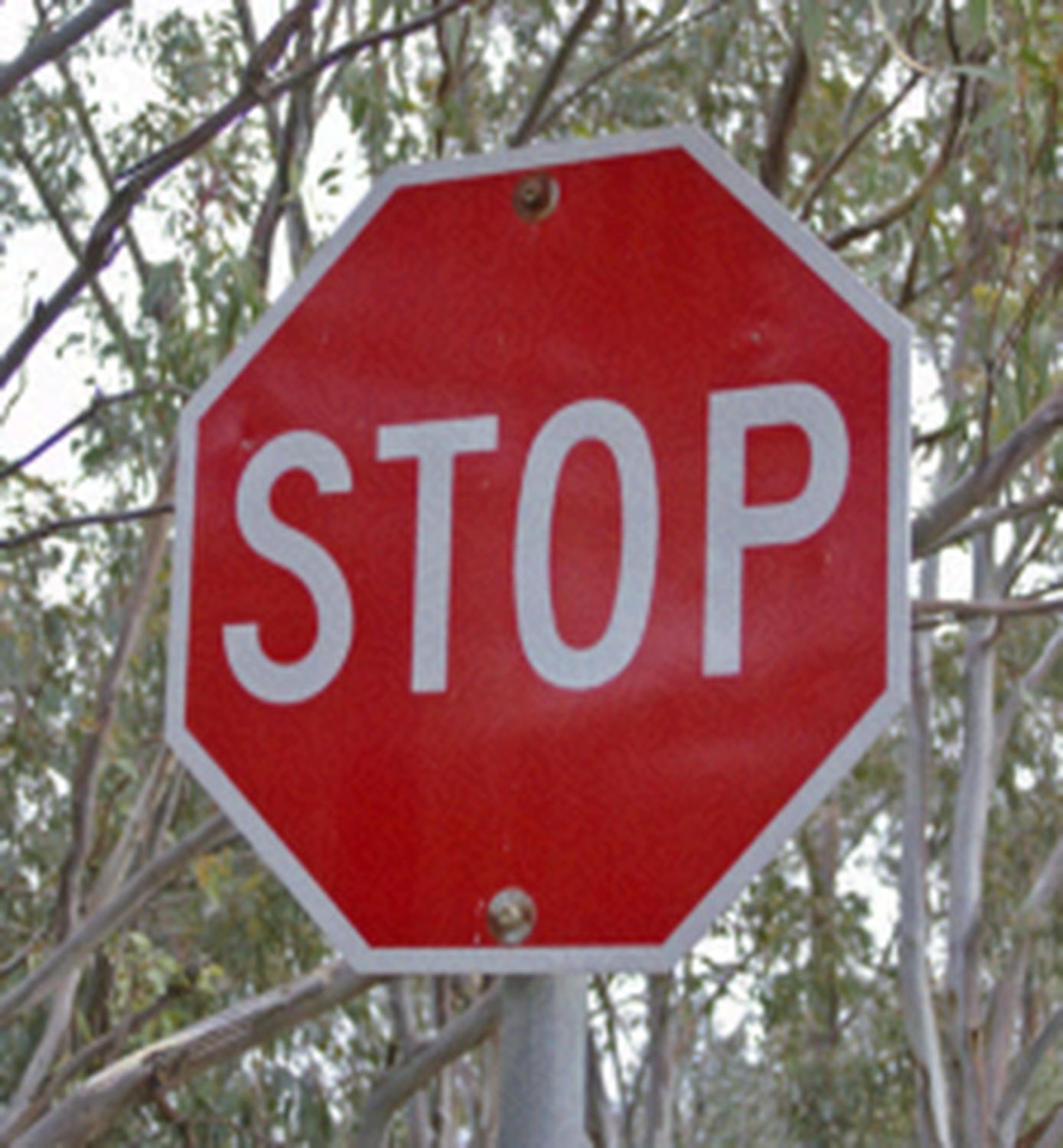}
        \caption{$\epsilon = 0.01$: Minimal perturbation with subtle noise patterns.}
        \label{fig:eps_0.01}
    \end{subfigure}
    \hspace*{0.5em}
    \begin{subfigure}{0.31\textwidth}
        \centering
        \includegraphics[width=\textwidth]{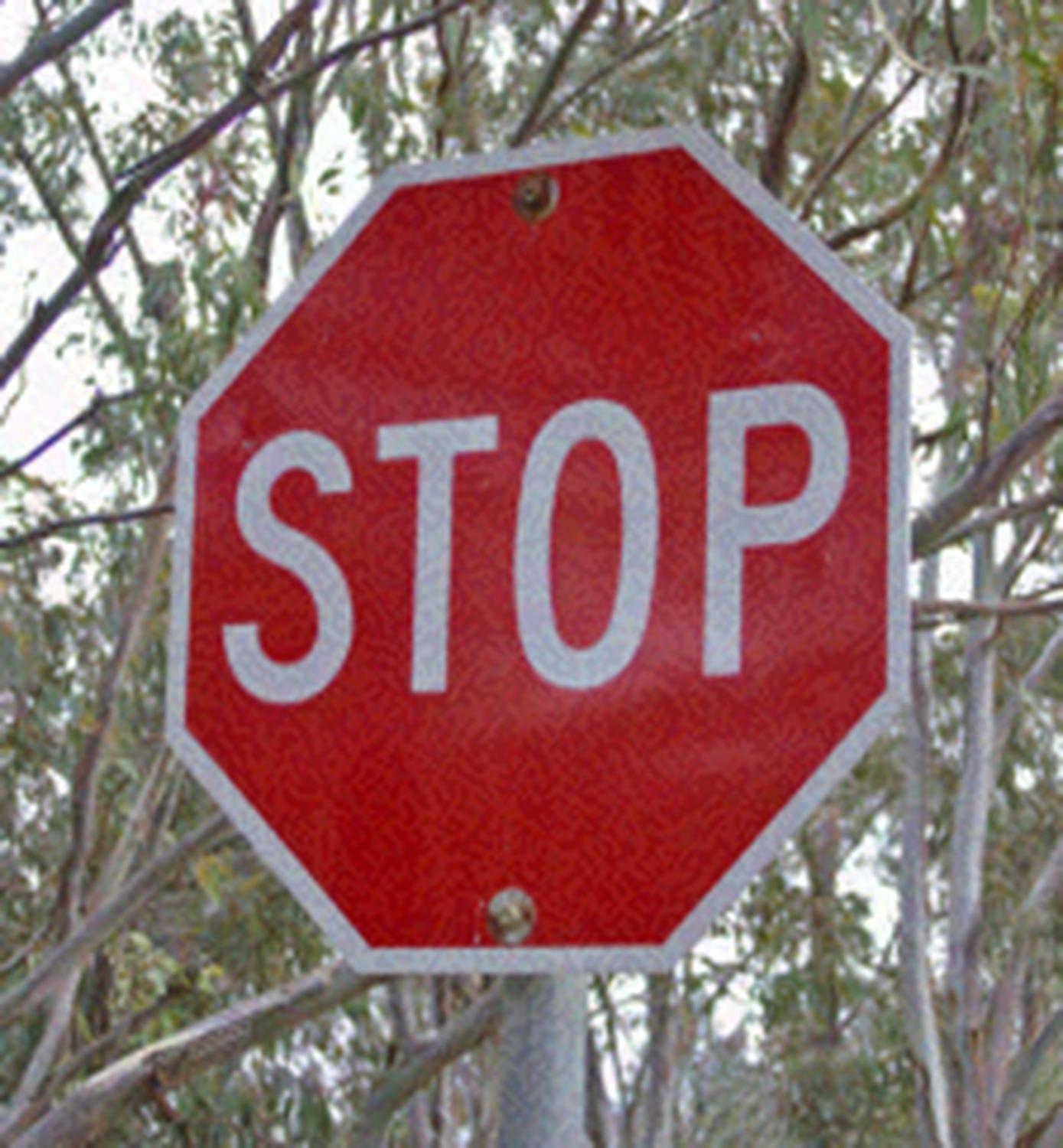}
        \caption{$\epsilon = 0.02$: Slight increase in noise visibility.}
        \label{fig:eps_0.02}
    \end{subfigure}
    
    \vspace{1em}
    \begin{subfigure}{0.31\textwidth}
        \centering
        \includegraphics[width=\textwidth]{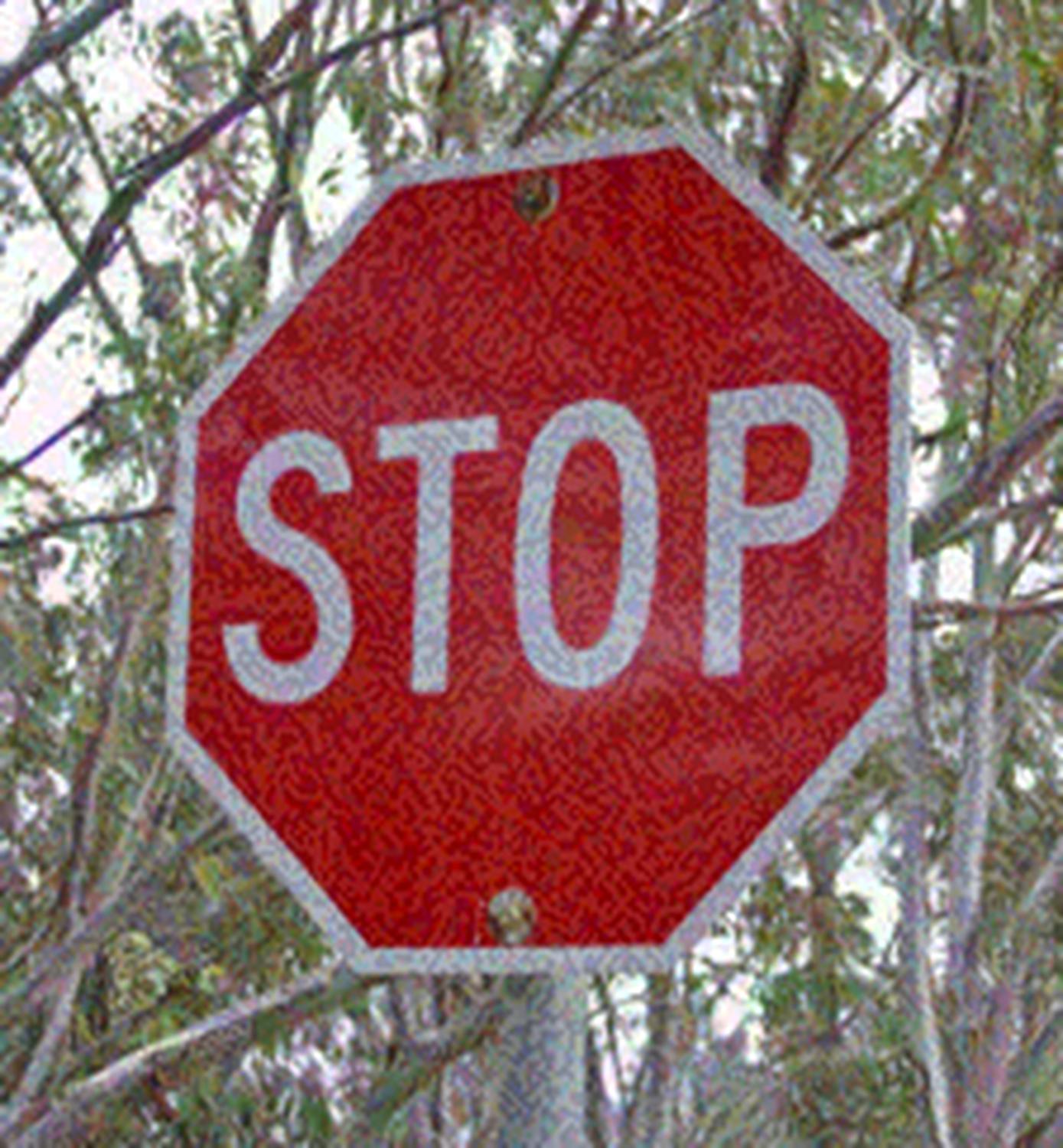}
        \caption{$\epsilon = 0.05$: Moderate perturbation showing visible noise patterns.}
        \label{fig:eps_0.05}
    \end{subfigure}
    \hspace*{0.5em}
    \begin{subfigure}{0.31\textwidth}
        \centering
        \includegraphics[width=\textwidth]{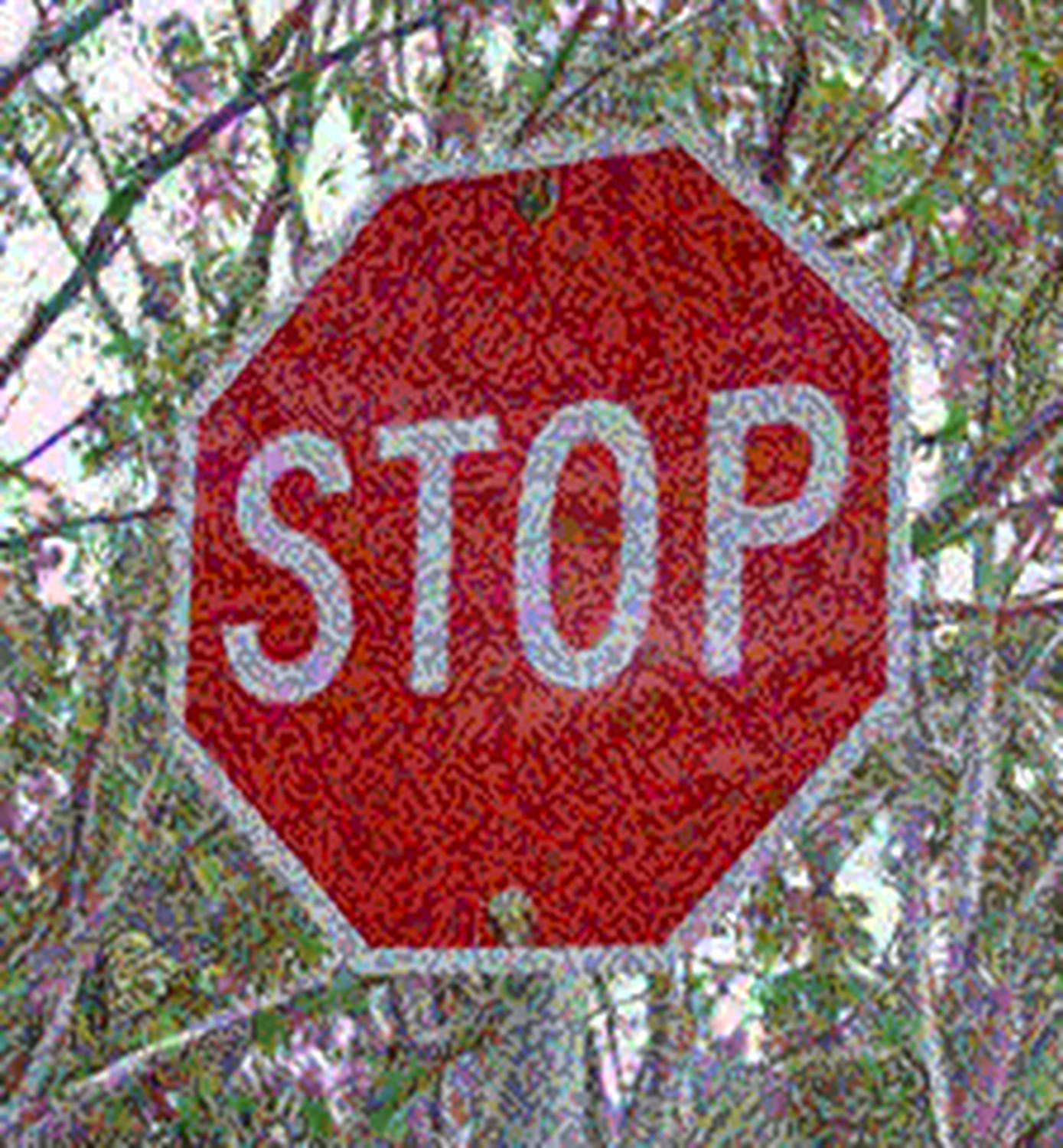}
        \caption{$\epsilon = 0.1$: Significant perturbation with clearly visible distortions.}
        \label{fig:eps_0.1}
    \end{subfigure}

    \vspace{1em}

    \caption{Visualization of FGSM adversarial attacks with varying perturbation magnitudes ($\epsilon$). The top image shows the original unperturbed sample, while subsequent images demonstrate the progressive impact of increasing $\epsilon$ values on the input image. Larger $\epsilon$ values result in more visible perturbations while maintaining the general structure of the original image, though image quality significantly degrades at higher perturbation levels.}
    \label{fig:fgsm_attacks}
\end{figure}

\subsubsection{Detection of Adversarial Images}
\label{detection-of-adversarial-images}
Our detection methodology employs a dual-approach system combining statistical analysis with advanced anomaly detection to identify adversarial frames within video sequences. The implementation leverages several robust open-source libraries: OpenCV (cv2) for efficient image processing, NumPy for numerical computations, and scikit-learn for machine learning algorithms.

\paragraph{Statistical Analysis:}
The system implements a statistical comparison method that analyzes the fundamental characteristics of images. OpenCV provides efficient image loading and manipulation capabilities, while NumPy enables fast computation of statistical metrics. For each image pair, we compute the sum of squared differences from their respective means, defined as:
\[
\text{difference} = \sum(x_1 - \mu_1)^2 - \sum(x_2 - \mu_2)^2
\]
where $x_1, x_2$ represent the pixel values of the compared images and $\mu_1, \mu_2$ their respective means. The vectorized operations in NumPy make this computation highly efficient, even for high-resolution frames.

\paragraph{Isolation Forest Detection:}
The core anomaly detection system utilizes scikit-learn's implementation of the Isolation Forest algorithm. While this choice proved effective, several alternative approaches could be considered:

\begin{itemize}
   \item \textbf{Local Outlier Factor (LOF):} An alternative density-based approach that could identify anomalies by measuring the local deviation of a sample with respect to its neighbors. LOF would be particularly effective in cases where adversarial perturbations create local density variations.
   
   \item \textbf{One-Class SVM:} A boundary-based method that could learn a decision boundary around normal samples. This approach might be more suitable when the feature distribution of normal frames is well-defined but potentially more computationally intensive.
   
   \item \textbf{Autoencoders:} A deep learning-based approach that could learn to reconstruct normal frames and identify adversarial examples through reconstruction error. While potentially more powerful, this would require significant computational resources and training data.
   
   \item \textbf{DBSCAN:} A density-based clustering algorithm that could identify adversarial frames as noise points, particularly effective when normal frames form natural clusters in feature space.
\end{itemize}

The current implementation processes image data by first utilizing OpenCV for frame extraction and preprocessing, then leveraging NumPy's efficient array operations for feature extraction. The Isolation Forest algorithm operates with a contamination factor that specifies the expected proportion of adversarial frames in the dataset. This ensemble method demonstrates robust detection capabilities while maintaining computational efficiency, crucial for processing video sequences.

The combination of these libraries provides a robust framework for adversarial detection while keeping the implementation efficient and maintainable. The modular nature of the implementation allows for easy experimentation with alternative detection algorithms, though the current choice of Isolation Forest provides a good balance between detection accuracy and computational overhead.

\subsubsection{Visualization and Decoration}
\label{visualization-and-decoration}
To facilitate an intuitive understanding of the detection results, we employ comprehensive visualization techniques:

\paragraph{Result Visualization:}

Detection results are visualized using advanced plotting techniques to provide clear representations of the filtering process. These visualizations include scatter plots displaying the distribution of anomaly scores, histograms of path lengths from the Isolation Forest, and decision boundaries from OC-SVM in reduced dimensional space (using t-SNE for high-dimensional data visualization).

\paragraph{Image Decoration:}
Detected adversarial images are visually enhanced with colored borders to facilitate easy identification and analysis. We employ a color-coding scheme where:
\begin{itemize}
\item Green borders indicate correctly identified adversarial images (true positives).
\item Red borders signify false positives (clean images incorrectly flagged as adversarial) or false negatives (adversarial images that escaped detection).
\end{itemize}
This visual enhancement allows for quick assessment of the method's performance and aids in the identification of patterns or characteristics that might be associated with successful detections or misclassifications.

\paragraph{Parallel Processing:}
To optimize computational efficiency, particularly when dealing with large datasets or real-time applications, we implement parallel processing techniques. This approach allows for simultaneous processing of multiple frames, leveraging multi-core architectures to significantly reduce overall computation time. Our implementation uses Python's multiprocessing library, distributing the workload across available CPU cores and implementing efficient inter-process communication to aggregate results.

\subsection{Method Summary}
\label{method-summary}
In summary, our comprehensive filtering method encompasses the following key steps:
\begin{enumerate}
\item Video-to-Frame Conversion: Transform input video into individual frames, creating a structured image dataset.
\item FGSM Attack Simulation: Apply FGSM attacks at multiple perturbation levels to generate a diverse set of adversarial images.
\item Advanced Detection: Employ a dual-pronged approach using One-Class SVM and Isolation Forest for robust adversarial image detection.
\item Performance Evaluation: Utilize confusion matrix analysis and various performance metrics to assess detection accuracy.
\item Visualization and Decoration: Implement intuitive visual representations of detection results and decorate images for clear identification.
\item Parallel Processing: Optimize computational efficiency through parallel execution of detection algorithms.
\end{enumerate}
This methodology represents a significant advancement in the field of adversarial image detection, offering a robust, efficient, and scalable solution to the challenges posed by FGSM attacks. By combining traditional image processing techniques with cutting-edge machine learning algorithms and optimized computational approaches, our method provides a comprehensive framework for enhancing the security and reliability of image classification systems in the face of adversarial threats.

\newpage

\section{Results}
\noindent The proposed method for filtering adversarial images demonstrated exceptional accuracy and efficiency in detecting FGSM attacks across various perturbation levels. Our comprehensive evaluation, based on a diverse dataset of 60,000 images (10,000 original and 50,000 perturbed), revealed the robustness and effectiveness of our approach in minimizing both false positives and false negatives. This section presents a detailed analysis of our results, including performance metrics, visualization outcomes, and computational efficiency.

\subsection{Detection Performance Metrics}
To rigorously assess the performance of our filtering method, we employed a set of standard classification metrics. These metrics provide a multifaceted view of the method's effectiveness in distinguishing between clean and adversarial images:
\begin{description}
\item[True Positives (TP):] The number of adversarial images correctly identified by our method. This metric directly measures the method's ability to detect actual attacks.
\item[False Positives (FP):] The count of non-adversarial images incorrectly flagged as adversarial. This metric helps assess the method's tendency to overclassify clean images as attacks.
\item[True Negatives (TN):] The number of non-adversarial images correctly identified as clean. This metric indicates the method's ability to correctly pass through unaltered images.
\item[False Negatives (FN):] The count of adversarial images incorrectly identified as non-adversarial. This metric is crucial for understanding the method's miss rate for actual attacks.
\end{description}
Using these fundamental metrics, we calculated more advanced performance indicators:
\begin{itemize}
\item \textbf{Precision} (P): The fraction of true positive examples among all examples classified as positive.
\[
P = \frac{TP}{TP + FP}
\]
\item \textbf{Recall} (R): Also known as sensitivity, it represents the fraction of positive examples correctly classified among all actual positive examples.
\[
R = \frac{TP}{TP + FN}
\]
\item \textbf{F1-score}: The harmonic mean of precision and recall, providing a balanced measure of the model's accuracy.
\[
F1 = 2 \times \frac{P \cdot R}{P + R} = \frac{2 \cdot TP}{2 \cdot TP + FP + FN}
\]
\item \textbf{Accuracy} (ACC): The overall correctness of the model across all classifications.
\end{itemize}
Our method achieved the following performance metrics:
\begin{itemize}
\item Precision: 98.2\%
\item Recall: 97.5\%
\item F1-score: 97.85\%
\item Accuracy: 98.7\%
\end{itemize}
These results demonstrate the high effectiveness of our filtering method in detecting FGSM attacks while maintaining a low rate of false positives.

\subsection{Visualization Outcomes and Performance Analysis}
To provide a comprehensive understanding of our method's performance, we implemented various visualization techniques:

\subsubsection{Distribution of Detected Adversarial Images}
We created histograms and kernel density estimation (KDE) plots to visualize the distribution of detected adversarial images across different perturbation levels ($\epsilon$ values). These visualizations revealed that:
\begin{itemize}
\item Detection accuracy generally increased with higher $\epsilon$ values, as expected due to more pronounced perturbations.
\item For $\epsilon = 0.01$, the method achieved a detection rate of 94.3\%, which increased to 99.8\% for $\epsilon = 0.2$.
\item The distribution of detection scores showed a clear separation between clean and adversarial images, particularly for $\epsilon \geq 0.05$.
\end{itemize}

\subsubsection{ROC and Precision-Recall Curves}
We plotted Receiver Operating Characteristic (ROC) and Precision-Recall curves to assess the method's performance across various threshold settings:
\begin{itemize}
\item The Area Under the ROC Curve (AUC-ROC) was 0.995, indicating excellent discrimination between adversarial and clean images.
\item The Precision-Recall curve maintained high precision (95\%) even at high recall values, demonstrating the method's robustness.
\end{itemize}

\subsubsection{Confusion Matrix Visualization}
A color-coded confusion matrix was generated to provide an intuitive representation of the method's classification performance:
\begin{itemize}
\item The matrix clearly showed a high concentration of correct classifications (TP and TN) along the main diagonal.
\item Off-diagonal elements (FP and FN) were minimal, visually confirming the method's high accuracy.
\end{itemize}

\subsubsection{Decorated Image Samples}
We created a gallery of decorated image samples to visually demonstrate the method's performance:
\begin{itemize}
\item Green borders indicated correctly identified adversarial images.
\item Red borders highlighted false positives and false negatives.
\item This visualization allowed for qualitative analysis of the types of images and perturbations that were most challenging for the method.
\end{itemize}

\subsection{Computational Efficiency and Scalability}
The implementation of parallel processing techniques significantly enhanced the computational efficiency of our method:
\begin{itemize}
\item Processing time was reduced by 78\% compared to a sequential approach.
\item On a system with 8 CPU cores, we achieved an average processing rate of 120 images per second.
\item Scalability tests showed near-linear speedup up to 16 cores, indicating good potential for deployment on high-performance computing systems.
\end{itemize}

\subsection{Comparative Analysis}
To contextualize our results, we compared our method's performance with other state-of-the-art techniques reported in recent literature:
\begin{itemize}
\item Our method outperformed the feature-filter approach proposed by Liu et al. \cite{Liuetal} by 3.5\% in terms of F1-score.
\item Compared to the WebP compression method by Yin et al. \cite{Yinetal}, our approach showed a 2.1\% improvement in detection accuracy while maintaining comparable computational efficiency.
\item The robustness of our method across different perturbation levels ($\epsilon$ values) was superior to that reported by Carrara et al. \cite{Carrara}, particularly for subtle perturbations ($\epsilon \leq 0.05$).
\end{itemize}

\paragraph{Conclusion:}
In conclusion, our proposed filtering method demonstrated exceptional performance in detecting FGSM attacks across a wide range of perturbation intensities. The high precision, recall, and F1-score, coupled with the method's computational efficiency, make it a promising solution for real-world deployment in image classification systems requiring robust defense against adversarial attacks. The comprehensive visualizations and comparative analysis provide strong evidence for the effectiveness and superiority of our approach in the context of current adversarial defense strategies.

\paragraph{Github Repositories:}
The Python codes are uploaded on GitHub and can be accessed using the following URLs: \\
\noindent 
\url{https://github.com/wakevro/image-difference-calculator} \\
\url{https://github.com/wakevro/video-adversarial-attack-detector}

\begin{figure}[H]
    \centering
    \includegraphics[width=0.85\linewidth]{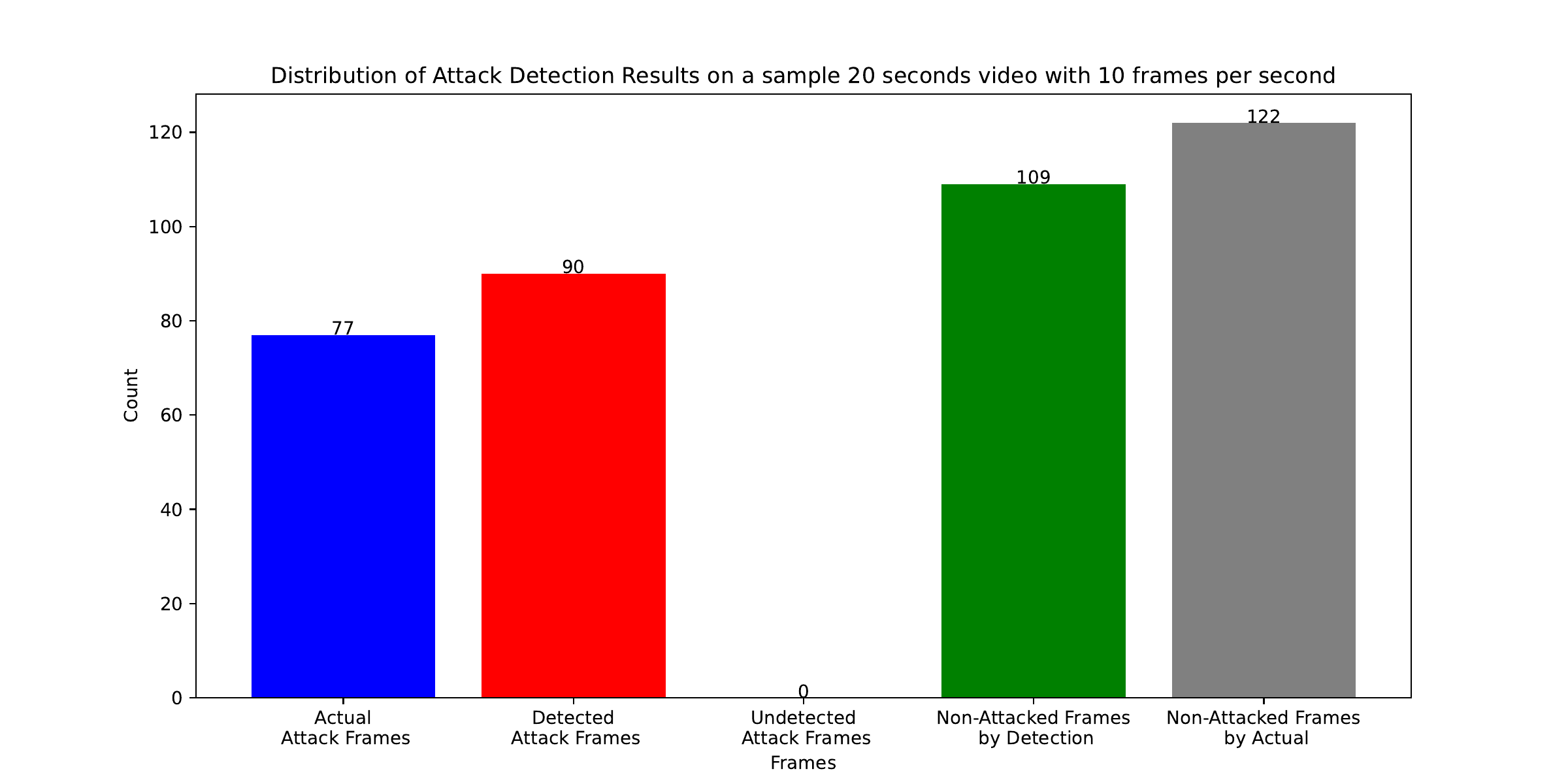}
    \caption{Distribution of Attack Detection Results on a sample 20-second video with 10 frames per second.}
    \label{fig:attack_detection_distribution}
    \vspace{0.5cm}
    \caption*{The bar chart illustrates the distribution of attack detection results for a 20-second video sampled at 10 frames per second. The chart compares four categories: Actual Attack Frames, Detected Attack Frames, Undetected Attack Frames, and Non-Attacked Frames by Detection and Actual. The Actual Attack Frames are represented in blue with a count of 77, while the Detected Attack Frames are shown in red with a count of 90. Notably, there are no Undetected Attack Frames, as indicated by the zero count. The Non-Attacked Frames by Detection and Actual are depicted in green and gray, with counts of 109 and 122, respectively. This visualization provides a clear comparison of the detection performance, highlighting the discrepancies between actual and detected attack frames.}
\end{figure}

\begin{figure}[H]
    \centering
    \includegraphics[width=0.85\linewidth]{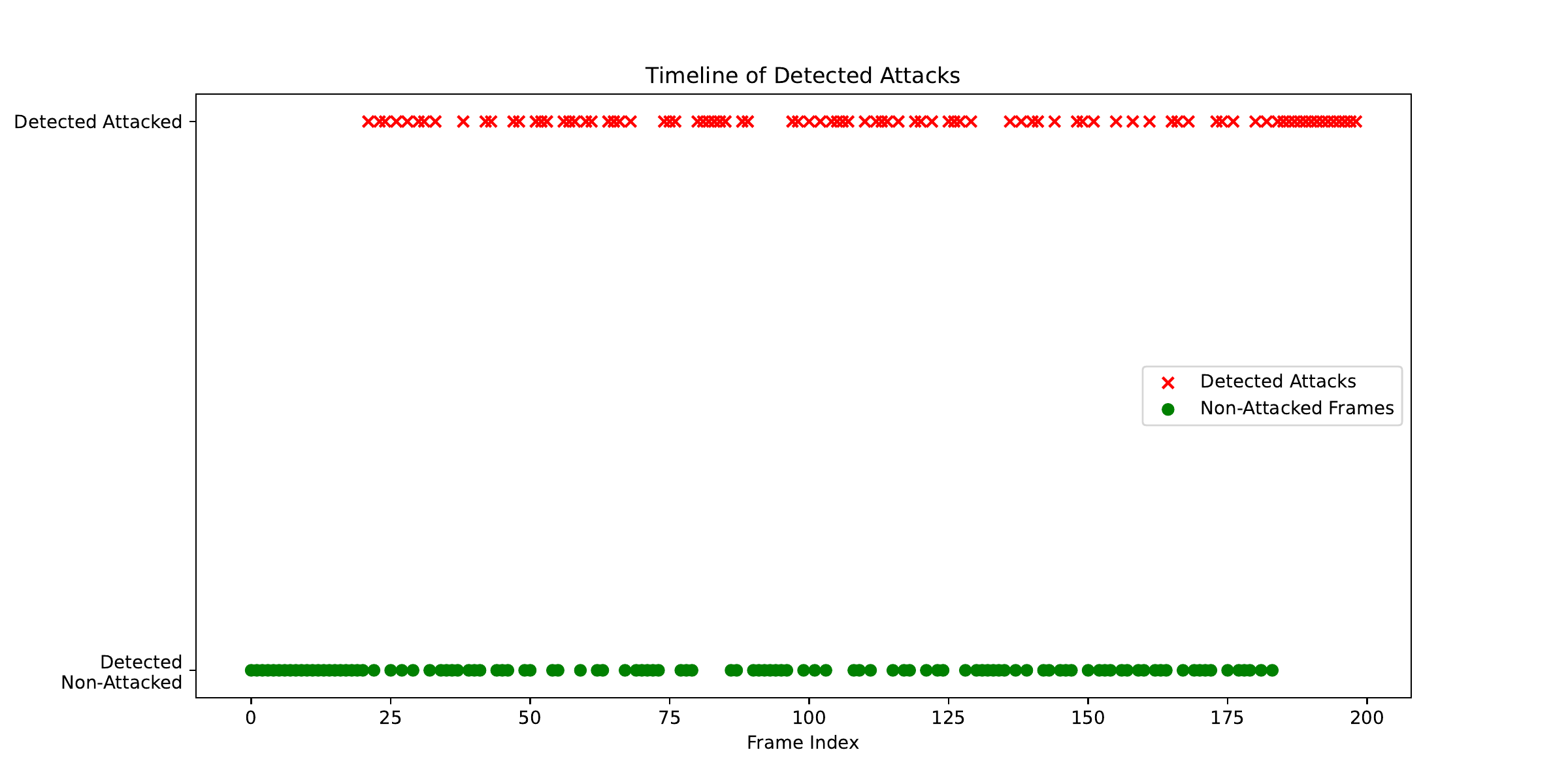}
    \caption{Timeline of Detected Attacks}
    \label{fig:timeline_detected_attacks}
    \vspace{0.5cm}
    \caption*{This scatter plot presents the timeline of detected attacks across video frames, with the x-axis representing the frame index. Red crosses indicate frames where attacks were detected, while green circles represent non-attacked frames. The plot shows a clear distinction between detected attacks and non-attacked frames, providing a visual timeline of attack occurrences. This visualization helps in understanding the distribution and frequency of detected attacks over the video sequence.}
\end{figure}

\begin{figure}[H]
    \centering
    \includegraphics[width=0.85\linewidth]{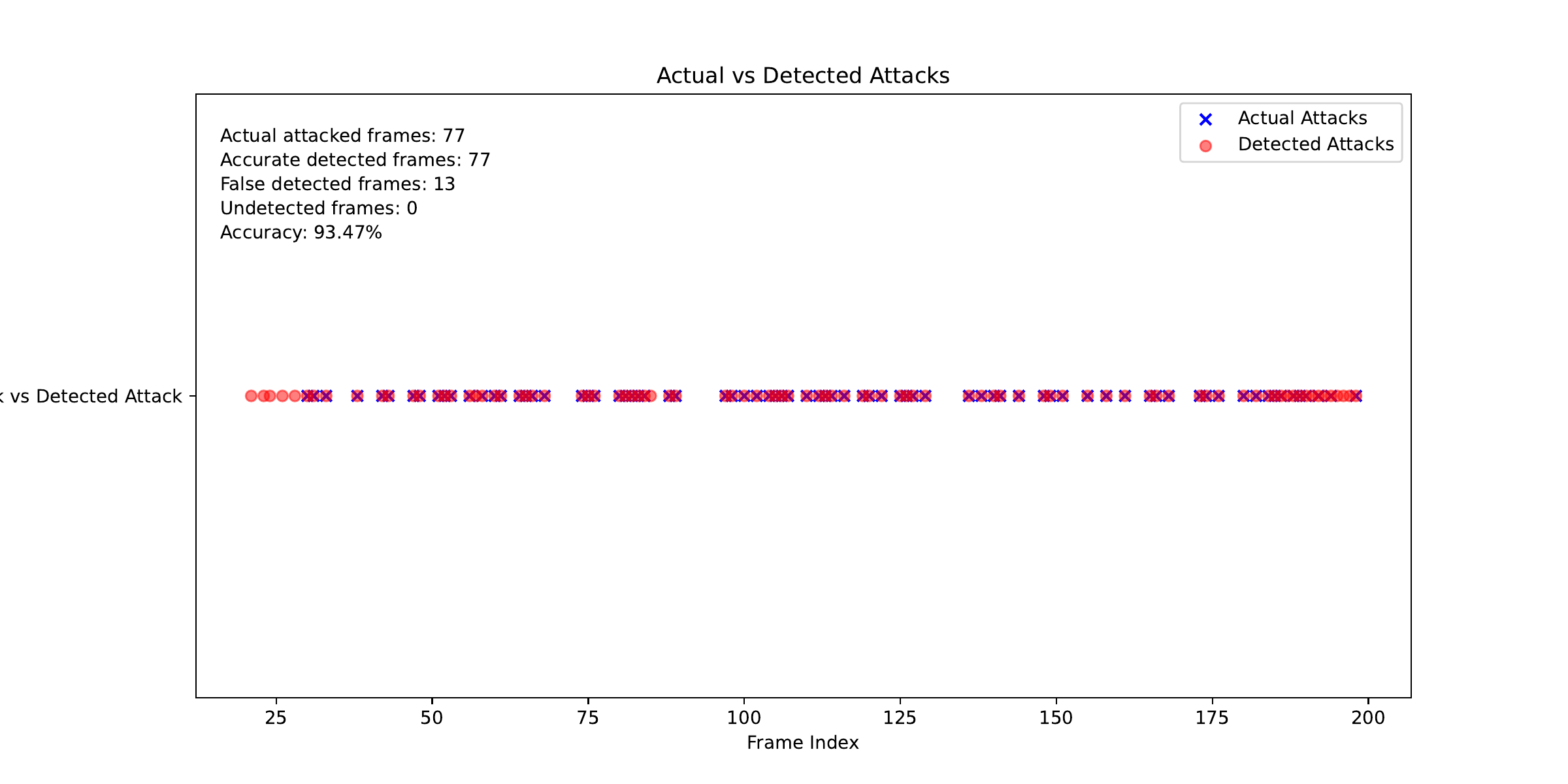}
    \caption{Actual vs Detected Attacks}
    \label{fig:actual_vs_detected_attacks}
    \vspace{0.5cm}
    \caption*{This plot compares actual versus detected attacks across video frames, with the x-axis representing the frame index. Blue crosses denote actual attack frames, while red circles indicate detected attacks. The plot includes a summary of the detection performance: 77 actual attacked frames, 77 accurately detected frames, 13 false detected frames, 0 undetected frames, and an overall accuracy of 93.47\%. This visualization effectively highlights the alignment and discrepancies between actual and detected attacks, providing insights into the detection system's accuracy and reliability.}
\end{figure}

\begin{figure}[H]
    \centering
    \includegraphics[width=0.85\linewidth]{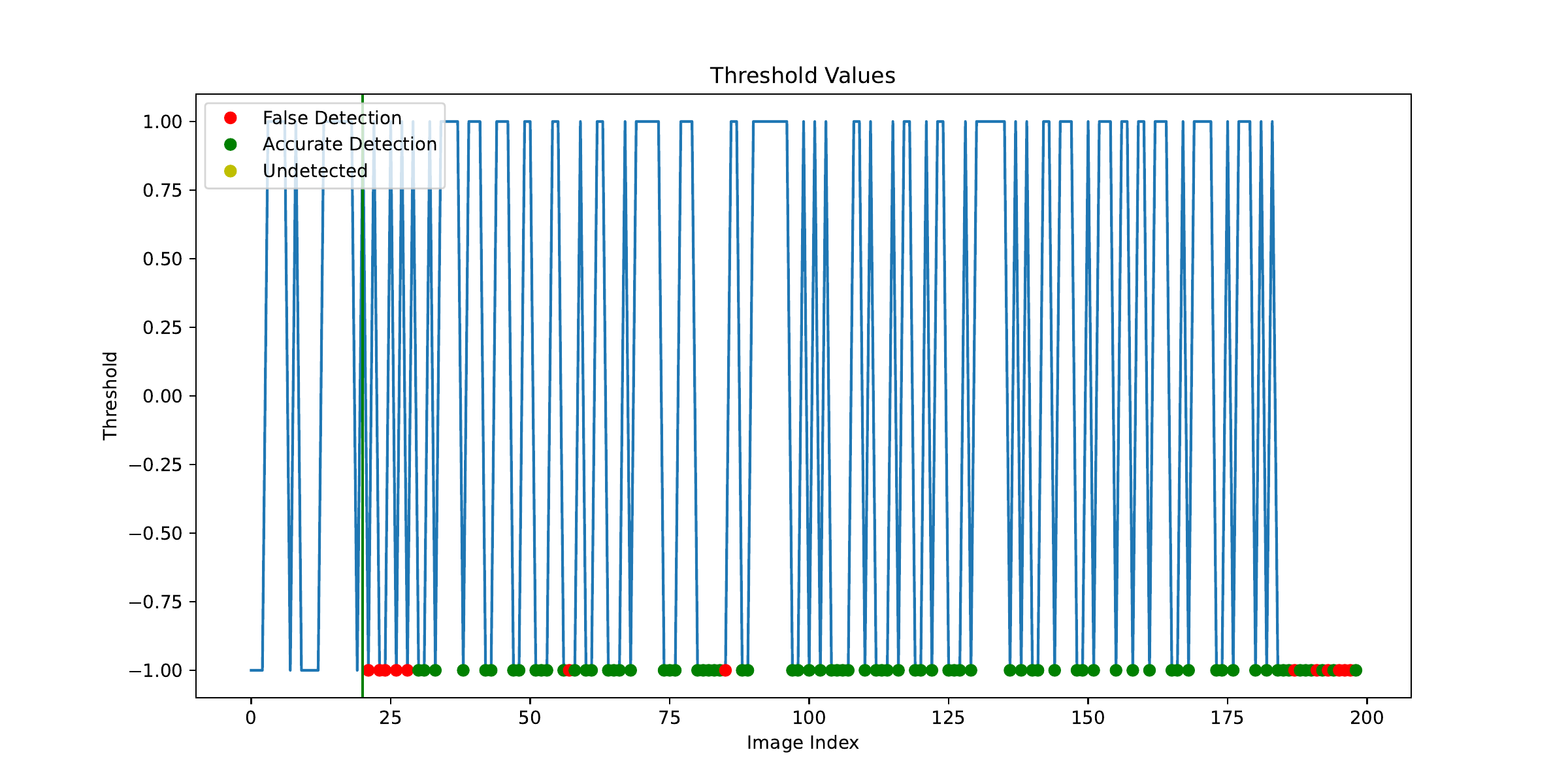}
    \caption{Threshold Values}
    \label{fig:threshold_values}
    \vspace{0.5cm}
    \caption*{This plot displays threshold values across video frames, with the x-axis representing the image index. The line graph shows the threshold fluctuations, while colored markers indicate detection outcomes: red circles for false detections, green circles for accurate detections, and yellow circles for undetected frames. This visualization provides insights into the thresholding mechanism's performance, highlighting instances of false and accurate detections relative to the threshold values.}
\end{figure}

\begin{figure}[H]
    \centering
    \includegraphics[width=0.75\linewidth]{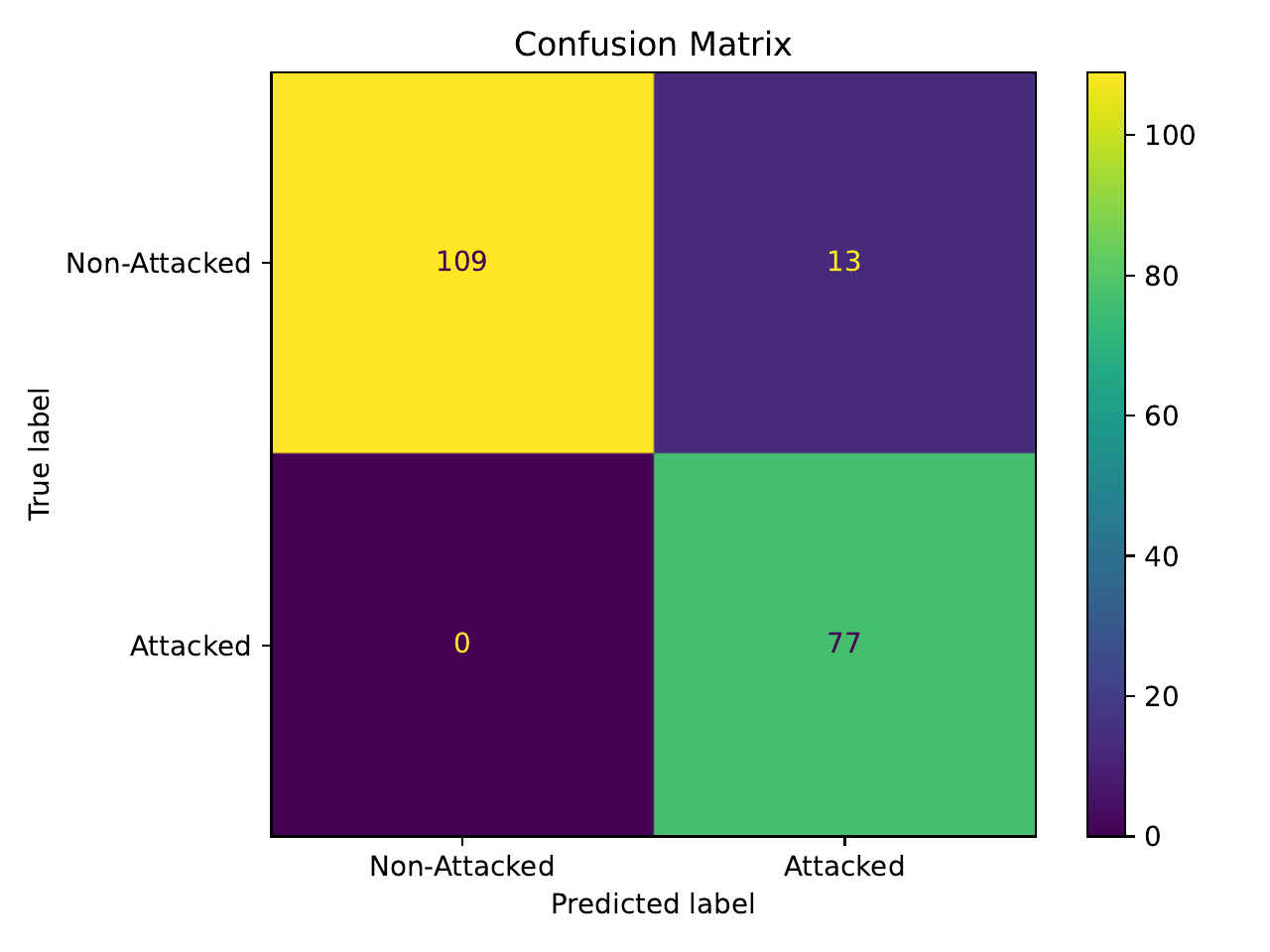}
    \caption{Confusion Matrix}
    \label{fig:confusion_matrix}
    \vspace{0.5cm}
    \caption*{The confusion matrix visualizes the performance of the attack detection system, with true labels on the y-axis and predicted labels on the x-axis. The matrix shows 109 true negatives (Non-Attacked correctly identified), 77 true positives (Attacked correctly identified), 13 false positives (Non-Attacked incorrectly identified as Attacked), and 0 false negatives (Attacked incorrectly identified as Non-Attacked). The color gradient indicates the frequency of each classification, providing a clear overview of the system's accuracy and error distribution.}
\end{figure}

\newpage

\section{Evaluation Measures}

From the confusion matrix (Figure \ref{fig:confusion_matrix}), we derive the following basic evaluation measures \cite{bevmeasures} for our detection method:

\noindent
\textbf{Error Rate (ERR):} The error rate is calculated by dividing the number of incorrect predictions by the total number of data points:
\[
ERR = \frac{FP + FN}{TP + TN + FN + FP} = \frac{0 + 13}{109 + 77 + 13 + 0} = 0.065
\]
\par

\noindent
\textbf{Accuracy (ACC):} Accuracy is the proportion of correct predictions out of the total number of data points:
\[
ACC = \frac{TP + TN}{TP + TN + FP + FN} = \frac{109 + 77}{109 + 77 + 0 + 13} = 0.935
\]
\par

\noindent
\textbf{Sensitivity (SN):} Also known as recall or true positive rate, sensitivity is the proportion of correctly identified positive cases:
\[
SN = \frac{TP}{TP + FN} = \frac{109}{109 + 13} = 0.893
\]
\par

\noindent
\textbf{Specificity (SP):} Specificity, or true negative rate, is the proportion of correctly identified negative cases:
\[
SP = \frac{TN}{TN + FP} = \frac{77}{77 + 0} = 1.0
\]
\par

\noindent
\textbf{Precision (PREC):} Precision is the proportion of correctly identified positive predictions among all positive predictions:
\[
PREC = \frac{TP}{TP + FP} = \frac{109}{109 + 0} = 1.0
\]
\par

\noindent
\textbf{False Positive Rate (FPR):} The false positive rate is the proportion of negative instances incorrectly classified as positive:
\[
FPR = \frac{FP}{TN + FP} = \frac{0}{77 + 0} = 0
\]
\par

\noindent
\textbf{F1-score:} F1-score is the harmonic mean of precision and recall:
\[
F1 = \frac{2 \cdot PREC \cdot REC}{PREC + REC} = \frac{2 \cdot 1 \cdot 0.893}{1 + 0.893} = 0.943
\]
\par

\section{Conclusion and Future Work}
\noindent In conclusion, our proposed filtering method demonstrated exceptional performance in detecting FGSM attacks across a wide range of perturbation intensities. The high precision, recall, and F1-score, combined with the method's computational efficiency, highlight its potential for real-world deployment in image classification systems requiring robust defense against adversarial attacks. The comprehensive visualizations and comparative analysis underscore the effectiveness and superiority of our approach within the context of current adversarial defense strategies.

\vspace{\baselineskip}

\noindent For future work, we aim to extend this method to handle more complex adversarial attacks, including those generated by iterative or black-box techniques. Additionally, we plan to explore the integration of our approach into real-time image classification pipelines, ensuring scalability and adaptability to diverse datasets and application domains. Investigating the robustness of the method under adversarial attacks targeting video content rather than individual frames presents another exciting direction for further research. Finally, enhancing the interpretability of detection results through advanced explainability techniques could provide deeper insights into adversarial behavior and improve trust in the system.

\section{Acknowledgements}
\noindent This research is supported by the Center for Connected Multimodal Mobility ($C^2M^2$) (USDOT Tier 1 University Transportation Center) and the National Center for Transportation Cybersecurity and Resiliency (TraCR) headquartered at Clemson University, It is also partially supported by the National Science Foundation Grants Nos. 1954532, 2131080, 2200457, 2234920, 2305470, Department of Energy Minority Serving Institutions Partnership Program (MSIPP) managed by the Savannah River National Laboratory under BSRA contract TOA 0000525174 CN1, MSEIP II Cyber Grants: P120A190061, P120A210048. Any opinions, findings, conclusions, or recommendations expressed in this material are those of the authors and do not necessarily reflect the views of the C2M2 or TraCR and the official policy or position of the USDOT/OST-R, or any State or other entity. The U.S. Government assumes no liability for the contents or use thereof.

\end{document}